\documentclass[11pt]{article}
\usepackage[margin=1in]{geometry}
\usepackage{times}
\usepackage{graphicx}
\usepackage{booktabs}
\usepackage{array}
\usepackage{longtable}
\usepackage{ragged2e}
\usepackage{caption}
\usepackage{authblk}
\usepackage[hidelinks]{hyperref}
\usepackage{enumitem}
\title{\vspace{-2em}\textbf{An ambiguity taxonomy for evaluating large language model performance on clinical registry abstraction: a multi-site 
prospective study}\vspace{-0.5em}}
\author[1]{James Matheson}
\author[1]{Betsy Castillo}
\author[2,*]{Andrew Y. Shin}
\author[2,3,*]{David Scheinker}
\affil[1]{Carta Healthcare, San Francisco, CA}
\affil[2]{Pediatrics, Stanford University School of Medicine, Stanford, CA}
\affil[3]{Medicine, Stanford University School of Medicine, Stanford, CA}
\affil[*]{Senior authors. Correspondence to: David Scheinker, \texttt{dscheink@stanford.edu}}
\date{}
\begin{document}
\maketitle
\begin{abstract}\noindent
\textbf{Objectives.} To evaluate large language model (LLM) performance on unprocessed electronic medical record (EMR) data for clinical registry 
abstraction.\par \noindent
\textbf{Methods.} We evaluated LLM performance answering registry questions for the American College of Cardiology National Cardiovascular Data 
Registry (ACC NCDR). In a pilot study at an academic medical center, the model identified candidate data sources for each registry question and 
experienced abstractors used these results to define question-specific document sets. In a validation study at a second center with a second ACC NCDR 
registry, the LLM answered questions using the question-specific document sets. Before reviewing any output, two abstractors independently 
established the ground truth and assigned each question to one of six categories, ordered by the ambiguity and clinical reasoning required to resolve 
it: Medication/Event Flag, Binary Clinical Presence, Administrative, Quantitative Laboratory/Physiologic, Clinical Interpretation, and Event 
Timing.\par \noindent
\textbf{Results.} The analytical sample comprised 9,430 abstractor answers reconciled to 4,715 consensus answers (501 pilot; 4,214 validation). In 
the pilot, candidate data sources per question averaged between 14.6 (SD 13.9) for demographics and 89.2 (SD 56.1) for history and risk factors. In 
validation, human inter-rater agreement was approximately 98\% while 87\% of LLM answers exactly matched consensus, 2\% partially, and 9\% did not. 
Mean question-level accuracy was 91.5\% (SD 13.4\%) across 157 questions with at least 20 answers, and declined as ambiguity increased, from 96\% for 
Medication/Event Flag to 62\% for Event Timing questions.\par \noindent
\textbf{Conclusions.} LLMs answering clinical registry questions on unprocessed EMR data achieved far lower accuracy than human abstractors, even 
with question-specific document scoping. LLM accuracy fell steadily as ambiguity and the level of required clinical reasoning increased.\par
\end{abstract}
\noindent\textbf{Keywords:} clinical natural language processing; large language model evaluation; inter-rater reliability; clinical data 
abstraction; registry quality; CathPCI; EPDI; multi-institutional study\vspace{1em}

\pagebreak

\section*{Introduction}
Large language models (LLMs) now perform at expert level on curated clinical vignettes and multiple-choice medical examinations 
\cite{Singhal2025,Goh2025}. Although these 
benchmarks have accelerated progress in clinical natural language processing, they differ substantially from real-world documentation tasks: most 
rely on narrowly defined questions with prespecified answers drawn from a single clinical context and summarize performance with a single overall 
metric, such as accuracy or macro-averaged F1 score \cite{Wornow2023,Bedi2025}. Such aggregates may obscure variation across clinically distinct tasks and may not reflect 
performance on unprocessed electronic medical record (EMR) data.

Two recent studies illustrate the gap: four leading LLMs generating ICD-9, ICD-10, and CPT codes at a single academic medical center all fell below 50\% exact-match accuracy against clinician-assigned codes \cite{Soroush2024}, and in a Fast Healthcare Interoperability Resources (FHIR)-compliant virtual EMR environment with 300 clinically derived tasks, the best model achieved a 69.7\% success rate with substantial variation across task categories \cite{Jiang2025}. 
The first study covered a single institution and task; the second's tasks remain structured and its environment artificial. Both point to the need to evaluate LLMs on real, unprocessed EMR documents, with their fragmented, redundant, and occasionally conflicting information.

Clinical registries provide a practical framework for such evaluation. A growing body of work applies LLMs to registry-style abstraction in oncology \cite{Enikeev2026}, pulmonary embolism \cite{Alwakeel2026}, and cardiovascular report classification \cite{vanderLoo2026}, typically reporting strong aggregate accuracy without explaining how and 
why performance varies across question types within a single registry. The American College of Cardiology National Cardiovascular Data Registry (ACC NCDR) defines hundreds of structured questions that trained hospital abstractors extract directly from the EMR for quality reporting, research, and regulatory submission \cite{Brindis2001}; these data have underpinned landmark national assessments of cardiovascular care \cite{Chan2011}. The questions span a wide spectrum 
of clinical reasoning, from transcribing a date of birth to synthesizing notes, laboratory results, and narratives into coded answers that may 
reasonably vary between trained abstractors; registries manage this variability through dual independent abstraction, reconciliation of 
disagreements, and measurement of inter-rater reliability (IRR), yielding an adjudicated reference standard well suited to evaluating LLMs on 
real-world, multi-source EMR tasks.

In this study we separated two sources of ambiguity as we measured LLM accuracy on unprocessed EMR data across categories of registry questions that demand different levels of clinical reasoning. Documentation Ambiguity arises across EMR sources, where the same information may appear in multiple notes that may be inconsistent or contradictory; we measure it by tasking the LLM with finding all data sources potentially relevant to each question and identifying whether it can confidently answer from each. Clinical Ambiguity arises from the question itself, which may permit several defensible interpretations even when the documentation is clear; we measure it with an ambiguity taxonomy in which experienced abstractors categorize questions, before the study, by the required level of clinical reasoning. We call the relationship between a category's position in the taxonomy and the model's 
accuracy the ambiguity--performance gradient.

\section*{Methods}
\subsection*{Study Design}
We conducted the study in two phases. The pilot phase, at Institution A, used the ACC NCDR Electrophysiology Device Implant registry (EPDI) and 
included 168 registry questions extracted from 3 de-identified patient records (501 question-patient observations). The validation phase, at 
Institution B, used the ACC NCDR Cardiac Catheterization and Percutaneous Coronary Intervention registry (CathPCI v5.7.1) and included 232 registry 
questions across 25 de-identified patient records (4,214 question-patient observations). The two institutions use different EMR systems and 
documentation conventions, so the same concept may appear in different note types, under different labels, with varying structure.
\subsection*{Reference Standard}
At each site, two practicing cardiac registry abstractors, credentialed through the ACC/NCDR certification program, independently answered each 
question; disagreements were reconciled through discussion to a single consensus answer, which served as the reference standard. Input data consisted 
of unprocessed EMR documentation, including clinical, procedure, consultation, and nursing notes, diagnostic imaging reports, discharge and transfer 
summaries, and available structured data elements.
\subsection*{Question Category Classification}
In the EPDI pilot, questions were grouped by registry section, a coarser scheme suited to the pilot's document-routing objective. Before the 
validation study, the abstractors independently grouped each registry question by the level of ambiguity, and corresponding clinical reasoning, its 
answer demands, assigning each to one of six categories ordered from lowest to highest reasoning demand: Medication/Event Flag (n = 62), Binary 
Clinical Presence (n = 41), Administrative (Transcription) (n = 18), Quantitative Laboratory/Physiologic (n = 10), Clinical Interpretation (n = 22), 
and Event Timing (n = 4); full definitions are in the Supplementary Methods. These counts reflect the 157 questions with at least 20 evaluable 
observations that constitute the analytic set (Table 3); all 232 validation questions were classified into the same six categories. A second reviewer 
independently classified a random 10\% sample of questions; agreement was 92\% (Cohen's $\kappa$ = 0.89), and disagreements were resolved by 
consensus.
\subsection*{LLM Protocol and Document-Context Targeting}
The same model configuration was used in both phases: Claude Sonnet 4.6 for more complex registry questions and Claude Haiku 3.5 for simpler 
questions, accessed via application programming interface within the institution's secure, HIPAA-compliant cloud environment; no protected health 
information left institutional boundaries. Each prompt contained the verbatim ACC NCDR question definition, including value sets where specified, and 
a standing instruction to answer when confident and decline when not; for questions with defined value sets, the model selected from the allowed 
options or explicitly abstained (full prompt structure in the Supplementary Methods).

In the pilot phase, we prompted the LLM to identify all documentation sources within each record that could plausibly contain the information needed 
for each question and, for each source, whether it could use it to answer. Across sequential iterations, these findings were developed into a 
structured mapping between question type and documentation context. In the validation phase, in order to isolate the effect of question ambiguity, 
the model received only the documents in each question's assigned context and produced a single answer for every patient record to which the question 
applied. Answers were scored against the adjudicated consensus as exact matches, partial matches (the answer contained the reference value), 
mismatches, or declinations.
\subsection*{Outcomes and Statistical Analysis}
The primary outcome was overall accuracy, the proportion of non-abstained answers achieving an exact or partial match. Secondary outcomes were mean 
question-level accuracy among the 157 questions with at least 20 answers (some questions do not apply to all patients) and category-level accuracy, 
the mean question-level accuracy within each category, with sensitivity tested by repeating the analysis at a 10-answer threshold. Differences across 
categories were tested with a Kruskal-Wallis H test and pre-specified pairwise Mann-Whitney U tests with rank-biserial effect sizes. The pilot, with 
3 patients, is reported descriptively; to characterize what drives answerability, we fit linear models of percent answerable on the number of 
candidate sources, with and without registry section as a categorical term (software details in the Supplementary Methods).

\section*{Results}
\subsection*{Pilot Study: Identifying Question-Specific Document Subsets}
The demographics section had the lowest average number of candidate documentation sources per question, 14.6 (SD 13.9), and history and risk factors the highest, 89.2 (SD 56.1) (Table 1). Record size, documentation density, and candidate-source counts differed substantially across patients (Supplementary Table S1). The answerable rate, the proportion of identified sources from which the model could generate an answer, was lowest for Episode of Care questions, 40.3\% (SD 28.5\%), and highest for laboratory questions, 79.0\% (SD 18.5\%) (Table 1).

Although the model identified many potentially relevant sources per question, a large fraction could not be used to answer; in several sections fewer than half of the candidate sources supported an answer (Table 1). A linear model of the association between the answerable rate and the number of candidate sources and the registry section explained 31\% of the variation (adjusted R\textsuperscript{2} = 0.31) and found that section membership was strongly associated with answerability (F = 18.3; p $<$ 0.0001) while the number of sources was not. This is reported descriptively given the three-patient pilot (Supplementary Figure S1).
These results were used to construct the document-context targeting scheme for the validation study: for each group of questions, the abstractors identified the document set to which the LLM's attention was restricted (Table 2; full retrieval specification in Supplementary Table S2).

\subsection*{Validation Study: Overall Answer and Match Rates}
In the validation study, the model generated answers for 99\% of observations (4,171 of 4,214) and declined for 1\% (43). Across all observations, 87\% were exact matches to the adjudicated abstractor consensus, 2\% were partial matches, 9\% were mismatches, and 1\% were abstentions. The analytical sample comprised 9,430 answers generated by experienced clinical abstractors and reconciled to 4,715 consensus answers (501 pilot, 4,214 validation), which served as the reference standard for evaluating LLM performance. The two human abstractors agreed on approximately 98\% of questions before reconciliation; the remainder, plus a few questions flagged during quality review, were reconciled to a single consensus answer.

Weighted aggregate accuracy across all observations was 89.6\%. Mean question-level accuracy, counting each of the 232 questions equally, was 84.0\%, a difference of 5.6 percentage points. For the 157 questions with at least 20 evaluable observations, mean accuracy was 91.5\% (SD 13.4\%; median 96.0\%; range 12.0--100.0\%): 61 questions (39\%) had 100\% accuracy, 18 (11.5\%) fell below 80\%, and 6 (3.8\%) were at or below 60\% (Figure 1). Relaxing the threshold to at least 10 observations (165 questions) left the distribution essentially unchanged (mean 90.6\%; median 96.0\%; 22 questions, 13.3\%, below 80\%); the eight additional questions were sampled more sparsely and skewed slightly harder (Supplementary Table S3 gives the full ranked distribution).

Mean question-level accuracy fell steadily as category complexity rose, from 96.1\% for Medication/Event Flag questions to 62.0\% for Event Timing questions (Table 3), a gap of 34.1 percentage points. Cross-category differences were highly significant (Kruskal-Wallis p $<$ 0.0001). Pairwise contrasts confirmed the gradient: Medication/Event Flag versus Clinical Interpretation (Mann-Whitney p $<$ 0.0001; rank-biserial r = 0.81) and Binary Clinical Presence versus Clinical Interpretation (p $<$ 0.0001) showed large effects, while the two most complex categories did not differ from each 
other (p = 0.39).

\subsection*{Accuracy in Clinically Ambiguous Questions}
Clinical Interpretation and Event Timing questions were over-represented among questions with accuracy below 80\%: the two categories represent 17\% of the analytic set (26 of 157 questions) but accounted for 12 of the 18 such questions (67\%), whereas among the 114 questions above 90\% accuracy only 6 (5\%) came from these categories, an approximately 13-fold enrichment (Figure 2).

Questions with perfect accuracy included binary demographics and comorbidities such as date of birth, diabetes, hypertension, and sex (all 100\% accuracy), along with most medication-administration and post-procedure event flags (Table 4); one notable exception was history of myocardial infarction, a Binary Clinical Presence question on which models achieved 68\% accuracy. Of the six questions at or below 60\%, five were interpretive or timing questions: cardiac cath lab visit indication (12\%), arrival date/time (20\%), the CSHA frailty scale and cardiovascular instability (both 56\%), and discharge date/time (60\%); the sixth was an administrative name question (48\%).

\subsection*{Abstention and Precision}
The model's rate of declining to answer rose with category complexity, from 0.2\% on Medication/Event Flag questions to 4.9\% on Event Timing questions. However, on questions with accuracy below 80\%, recall remained above 90\% while precision fell below 60\%. 

\section*{Discussion}
In the validation phase of the evaluation of frontier LLMs across two ACC/NCDR registries at two academic medical centers, overall LLM accuracy was 89.6\% on 4,214 questions, falling from 96.1\% on simple binary questions to 62.0\% on questions requiring the most clinical reasoning to resolve ambiguity. The two most complex categories, Clinical Interpretation and Event Timing, were about thirteen times over-represented among the questions with the lowest accuracy. Although the model was instructed to abstain when unsure, it seldom did including on questions on which it achieved very low accuracy.

\subsection*{Unprocessed EMR data}
Models evaluated on clinical vignettes are asked a clean question with one adjudicated answer from a single clinical context. The medical record is not clean: the same fact can sit in a progress note, a discharge summary, and a nursing flowsheet, recorded differently under different labels across encounters months apart. In our pilot, a single history or risk-factor question drew on an average of 89 candidate sources per patient, fewer than half of which the model judged usable. These findings align with recent reports that feeding an LLM more of the chart made its diagnoses worse, because models are sensitive to both the amount and order of information \cite{Hager2024}. Because the pilot scoped each question to a defined document set, the validation phase could examine performance as a function of complexity alone.

\subsection*{Implications for evaluating and interpreting LLM performance}
Our findings suggest that model performance should be reported by category and by question, rather than in aggregate. Our weighted accuracy exceeded our question-mean accuracy by 5.6 percentage points because the easy questions recur in many records, so improving on common, easy questions raised the weighted score more than improving on rare, clinically challenging questions. This matches, and adds clinical context to, concerns previously raised about aggregating across questions for conversational tasks \cite{OpenAI2025}. The performance gradient is not peculiar to our registries: earlier studies found near-perfect accuracy on left ventricular function but markedly lower accuracy on culprit vessel in the same angiography reports and concordance on thyroid pathology reports highest on simple binary and categorical questions and falling with textual interpretation \cite{vanderLoo2026,Lee2025}. The agreement between model performance and the ambiguity taxonomy presented here suggests that it may serve as a useful framework for evaluation, and 
that other schemes should preserve the divide between direct extraction and interpretive synthesis.

The way the model erred is itself informative. On the hard questions it answered anyway, often incorrectly, the strategy that training on single 
labels graded by accuracy rewards, since producing the most probable label beats conceding ignorance \cite{Kalai2025}. A model trained to decline when unsure would instead let those questions be routed to a person. 

The models fared worst on Event Timing questions. In those, the right answer usually depends on reconciling several documents rather than reading any one, since a timestamp may differ across a triage note, an EMS handoff, a nursing intake, and a billing record. Work on verifying facts against the EMR calls this a needle-in-a-haystack problem, with evidence scattered through the chart and hard to localize even with retrieval \cite{Chung2026}. 

The ambiguity taxonomy is consistent with other recent work. Medication extraction, with end-to-end F1 scores of 0.69 and 0.82 in French and English \cite{Fabacher2025}, is a Medication/Event Flag task checkable against structured records. Systematic-review abstract screening, where LLM sensitivity approaches 1.0 \cite{Sanghera2025}, is a binary-presence question with a constrained answer set and a single self-contained source. Some studies find that the hard questions are hard because they are fundamentally ambiguous, in the sense that experts disagree about the answer \cite{OMalley2005,Reamaroon2019}. The data argue against this, though they do not settle it: if the model's 12\% to 20\% accuracy on the hardest questions reflected genuine disagreement about truth, the abstractors should have disagreed nearly as often, and they did not. We think it more likely that the documentation on hard questions is multi-source and sometimes contradictory, and reaching the answer takes reconciliation: the abstractors reconciled by thinking or 
talking the case through. The 98\% agreement suggests that this was possible for most questions, but it is an aggregate lifted by the many easy questions. Measuring category-specific and question-specific human IRR is an important area for future work.

A model graded only on a single gold label learns that guessing beats abstaining, which is one explanation for why it answered confidently on the questions it was most likely to get wrong \cite{Kalai2025}. Our findings align with a decade of work on training models to handle ambiguity and suggest a specific clinical application: allow models to measure their own uncertainty and index against human uncertainty, what we call IRR-indexed loss. Model IRR, or semantic entropy, is measured by sampling several answers to the same question, clustering them by meaning, and using the spread across clusters to estimate how unsure the model is.\cite{Farquhar2024} Currently, it is used primarily to flag potential hallucinations. Along with human IRR, it could be 
used to direct training and evaluation. When human IRR is high, the loss should push the model to spend more time on inference: read more carefully, reconcile across sources, and reason more deeply. When human IRR is low, the question is ambiguous and the fiction of a single correct answer will encourage confident guessing or hallucination; the target should be a distribution over the defensible answers.

\subsection*{Limitations}
The number of patients in both studies was relatively small (3 pilot, 25 validation patients), so per-question estimates are wide. For the human IRR, we rely on category-level averages; per-category IRR, a key follow-up, was not recorded. The reference standard reflects credentialed abstractors, so other registries and later models will need their own evaluation.

\section*{Conclusion}
Across two ACC/NCDR registries at two institutions with 98\% human IRR, LLM accuracy fell with question ambiguity, from 96\% on the simplest category 
to 62\% on the most complex. Studies of LLM performance on clinical abstraction tasks should be reported and interpreted across categories of 
different levels of ambiguity, rather than in aggregate.

\textbf{Acknowledgments:} The authors thank the clinical abstractors of Carta Healthcare for the dual abstraction and adjudication that produced the 
reference standard, the Carta engineering team for the evaluation infrastructure, and the participating institutions for access to the de-identified 
records used in this study.\par
\textbf{Funding:} No funding\par
\textbf{Conflict of interest:} Authors JM and BC are employed by Carta Healthcare, the company that conducted the study. Authors AS and DS advise 
Carta Healthcare.\par
\textbf{Data availability:} Question-level performance data and analysis code are available from the corresponding author upon reasonable request.\par
\clearpage
\begin{table}[h]\centering\small
\caption{Candidate documentation sources per question and answerable rate. Pilot study (Institution A, EPDI registry, n = 3 patients); values 
aggregated across patients by registry section.}
\begin{tabular}{@{}lccc@{}}\toprule
Registry Section & N Questions & Candidate Sources / Q (Mean $\pm$ SD) & Answerable \% (Mean $\pm$ SD)\\\midrule
A. Demographics & 9 & 14.6 $\pm$ 13.9 & 50.5\% $\pm$ 19.3\%\\
B. Episode of Care & 9 & 15.1 $\pm$ 14.2 & 40.3\% $\pm$ 28.5\%\\
C. History and Risk Factors & 66 & 89.2 $\pm$ 56.1 & 42.2\% $\pm$ 14.7\%\\
D. Diagnostic Studies & 13 & 48.4 $\pm$ 51.3 & 64.0\% $\pm$ 21.2\%\\
E. Labs & 3 & 25.1 $\pm$ 6.7 & 79.0\% $\pm$ 18.5\%\\
F. Procedure Information & 8 & 22.1 $\pm$ 24.8 & 50.3\% $\pm$ 17.1\%\\
G. Device Implant/Explant & 23 & 29.2 $\pm$ 27.8 & 60.3\% $\pm$ 21.2\%\\
H. Lead Assessment & 11 & 27.5 $\pm$ 26.4 & 57.7\% $\pm$ 18.4\%\\
I. Intra/Post-Procedure Events & 15 & 29.0 $\pm$ 18.3 & 47.2\% $\pm$ 14.8\%\\
J. Discharge & 10 & 12.6 $\pm$ 15.8 & 75.0\% $\pm$ 21.1\%\\
\bottomrule\end{tabular}\end{table}
\begin{figure}[h]\centering\includegraphics[width=0.85\textwidth]{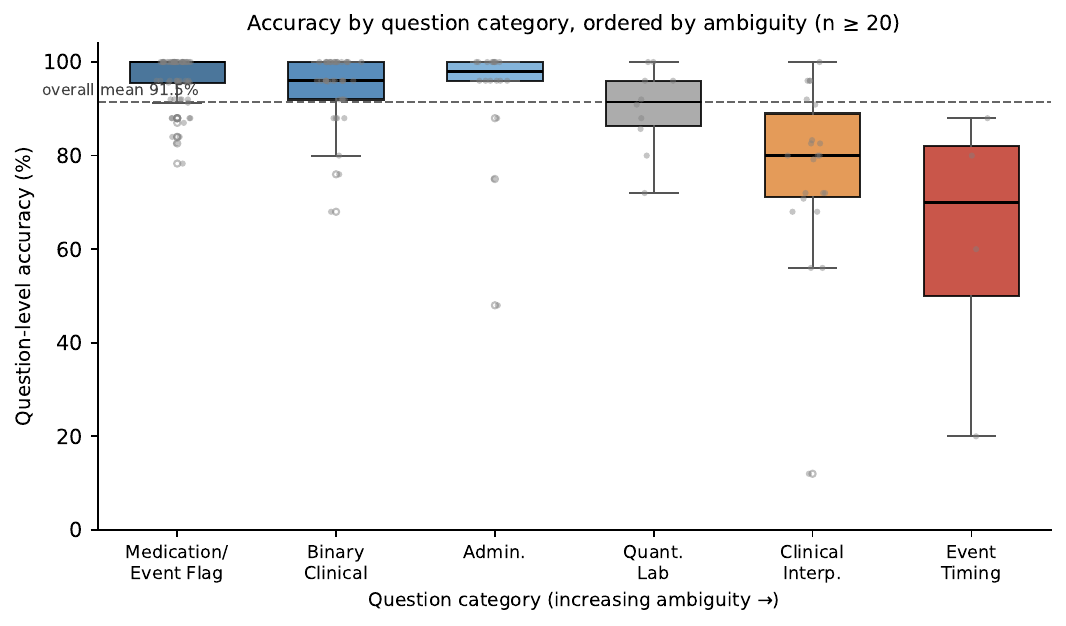}
\caption{Question-level accuracy by category, ordered left to right by increasing clinical ambiguity (157 CathPCI questions, n $\geq$ 20 
observations). Boxes show interquartile range and median; whiskers extend to 1.5$\times$IQR; points show individual questions. Dashed line marks the 
overall mean (91.5\%).}\end{figure}
\begin{table}[h]\centering\small
\caption{Document-context targeting derived from the pilot. Each row gives a set of registry questions, the documentation context to which the LLM 
was restricted, the EMR resources that context retrieves, and the number of CathPCI validation-study questions routed to it.}
\begin{tabular}{@{}p{3.6cm}p{2.4cm}p{4.2cm}c@{}}\toprule
Example Question & Registry section & EMR Resources Selected by Trained Abstractors & \# Q\\\midrule
Arrival Date/Time; Discharge Date/Time; Procedure Start Date/Time & All Documents & Full patient record, no filters & 3\\\addlinespace
Patient Last/First Name; DOB; Sex; Race; Hispanic Origin; Health Insurance; ZIP Code & Patient Demographics \& H\&P & Patient resource, Encounter, 
History \& Physical, Consult notes, Discharge summaries & 16\\\addlinespace
Admission Source; Prior MI; Hypertension; Diabetes; Dyslipidemia; NYHA Class; Prior PCI; Tobacco Use; Stress Test Type/Result & Broad Clinical Notes 
& Encounter, Progress notes, Discharge summaries, H\&P, Consult, Procedure, Nursing, Transfer summaries & 111\\\addlinespace
Hemoglobin; Sodium; Creatinine (pre/post); Potassium; BUN; Calcium Score Assessed & Lab Results & Lab Observations, Diagnostic reports (CBC, 
metabolic panels, cath-lab labs) & 12\\\addlinespace
Discharge Medication Code (RxNorm); Pre-procedure / Procedure Medication Administration; GDMT Maximum Dose & Medications & MedicationRequest, 
MedicationAdministration, MedicationStatement, Discharge summaries & 5\\\addlinespace
CABG Status; PCI Date; Discharge Status / Location / Hospice; Comfort Care; Cardiac Rehab Referral & Discharge Summary & Discharge summary documents 
only & 10\\\addlinespace
Prior LVEF Assessed; Most Recent LVEF \% / Date; ECG Results & Echocardiography \& Imaging & Echocardiography, cardiac ultrasound reports & 
4\\\addlinespace
\bottomrule\end{tabular}\end{table}
\begin{table}[h]\centering\small
\caption{Accuracy by category (CathPCI validation study, questions with at least 20 observations), ordered from simplest to most complex. Human 
inter-rater reliability was approximately 98\% in aggregate across the validation set; question-level IRR by category was not separately recorded.}
\begin{tabular}{@{}lccccc@{}}\toprule
Category & N & Mean (\%) & Median (\%) & SD (\%) & Range (\%)\\\midrule
Medication / Event Flag & 62 & 96.1 & 100.0 & 5.3 & 78.3--100.0\\
Binary Clinical Presence & 41 & 94.9 & 96.0 & 7.0 & 68.0--100.0\\
Administrative (Transcription) & 18 & 93.7 & 98.0 & 13.0 & 48.0--100.0\\
Quantitative Laboratory / Physiologic & 10 & 90.1 & 91.5 & 9.0 & 72.0--100.0\\
Clinical Interpretation & 22 & 76.6 & 80.0 & 19.0 & 12.0--100.0\\
Event Timing & 4 & 62.0 & 70.0 & 30.4 & 20.0--88.0\\
All questions (n $\geq$ 20) & 157 & 91.5 & 96.0 & 13.4 & 12.0--100.0\\
\bottomrule\end{tabular}\end{table}
\begin{table}[h]\centering\small
\caption{Selected questions with very high and very low accuracy, with interpretive annotations.}
\begin{tabular}{@{}p{3cm}cp{2.4cm}p{5cm}@{}}\toprule
Question & Acc. & Category & Interpretive note\\\midrule
Cath lab visit indication & 12\% & Clinical Interpretation & Integrates presenting symptoms, ECG, troponin trajectory, and clinical narrative to 
select from a multivalent coded value set.\\\addlinespace
Arrival date/time & 20\% & Event Timing & Timestamp spans triage note, EMS handoff, and nursing intake; the canonical moment is contested when 
sources conflict.\\\addlinespace
CSHA frailty scale & 56\% & Clinical Interpretation & Frailty gestalt not uniformly captured across note types; abstractor-dependent.\\\addlinespace
Cardiovascular instability & 56\% & Clinical Interpretation & Hemodynamic instability threshold varies by clinician and documentation 
style.\\\addlinespace
Discharge date/time & 60\% & Event Timing & Discharge timestamp may conflict across physician note, nursing note, and billing record.\\\addlinespace
History of myocardial infarction & 68\% & Binary Clinical (edge case) & Patient-reported vs.\ confirmed MI and Q waves of uncertain age elevate 
difficulty.\\\addlinespace
Hypertension & 100\% & Binary Clinical & Well-defined comorbidity with explicit ACC/NCDR coding; deterministic extraction from problem 
list.\\\addlinespace
Sex & 100\% & Administrative & Single discrete value from patient registration; no inference required.\\\addlinespace
Discharge medication administration & 91--100\% & Medication/Event & Binary presence of a coded medication at discharge; directly verifiable against 
structured orders.\\\addlinespace
\bottomrule\end{tabular}\end{table}
\begin{figure}[h]\centering\includegraphics[width=0.7\textwidth]{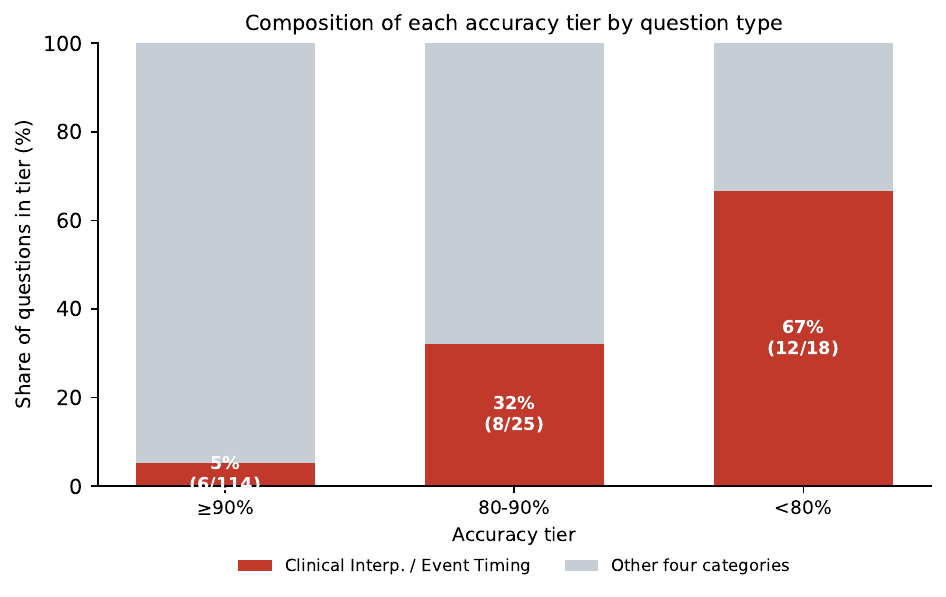}
\caption{Composition of each accuracy tier by question type. Bars show the share of questions in each tier classified as Clinical Interpretation or 
Event Timing (red) versus the other four categories (gray), with counts inset. These two categories make up 5\% of the $\geq$90\% tier but 67\% of 
the <80\% tier, an approximately 13-fold enrichment.}\end{figure}
\begin{figure}[h]\centering\includegraphics[width=0.75\textwidth]{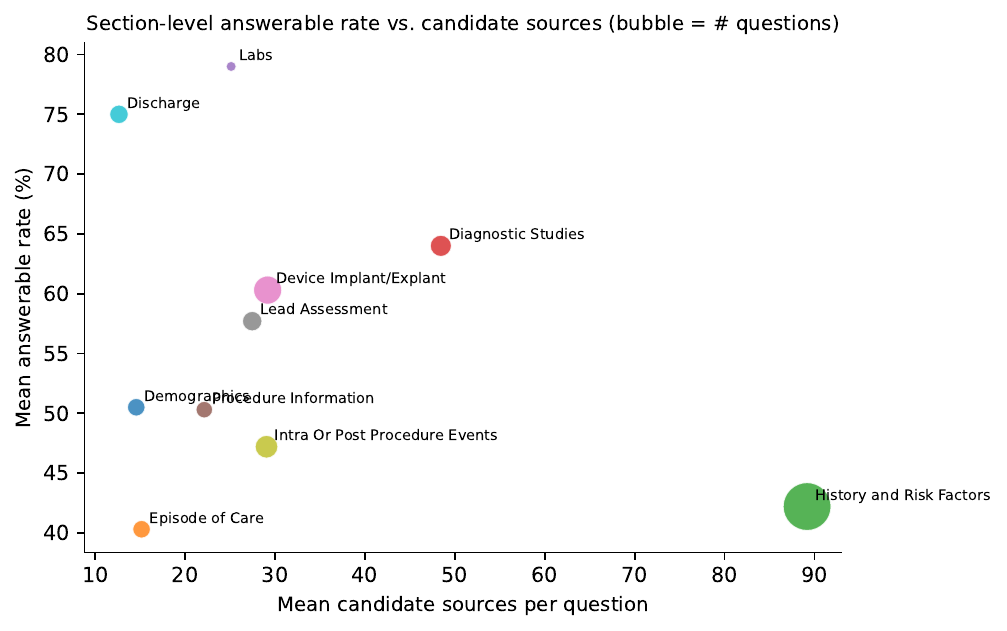}
\caption{\textbf{Supplementary Figure S1.} Section-level mean answerable rate versus mean candidate sources per question in the EPDI pilot (bubble 
area proportional to number of questions). Answerable rate varies by section but shows no consistent relationship with the number of candidate 
sources, indicating that answerability is driven by question type rather than source volume.}\end{figure}
\clearpage
\small
\bibliographystyle{unsrt}
\bibliography{references}

@article{Singhal2025,
  author  = {Singhal, K. and Tu, T. and Gottweis, J. and others},
  title   = {Toward expert-level medical question answering with large language models},
  journal = {Nat Med},
  year    = {2025},
  volume  = {31},
  number  = {3},
  pages   = {943--950},
  doi     = {10.1038/s41591-024-03423-7},
  note    = {doi:10.1038/s41591-024-03423-7}
}

@article{Goh2025,
  author  = {Goh, E. and Gallo, R. J. and Strong, E. and others},
  title   = {{GPT-4} assistance for improvement of physician performance on patient care tasks: a randomized controlled trial},
  journal = {Nat Med},
  year    = {2025},
  volume  = {31},
  number  = {4},
  pages   = {1233--1238},
  doi     = {10.1038/s41591-024-03456-y},
  note    = {doi:10.1038/s41591-024-03456-y}
}

@article{Wornow2023,
  author  = {Wornow, M. and Xu, Y. and Lehman, E. P. and others},
  title   = {The shaky foundations of large language models and foundation models for electronic health records},
  journal = {NPJ Digit Med},
  year    = {2023},
  volume  = {6},
  number  = {1},
  pages   = {135}
}

@article{Bedi2025,
  author  = {Bedi, S. and Liu, Y. and Orr-Ewing, L. and others},
  title   = {Testing and evaluation of health care applications of large language models: a systematic review},
  journal = {JAMA},
  year    = {2025},
  volume  = {333},
  number  = {4},
  pages   = {319--328},
  doi     = {10.1001/jama.2024.21700},
  note    = {doi:10.1001/jama.2024.21700}
}

@article{Soroush2024,
  author  = {Soroush, A. and Glicksberg, B. S. and Zimlichman, E. and others},
  title   = {Large language models are poor medical coders: benchmarking of medical code querying},
  journal = {NEJM AI},
  year    = {2024},
  volume  = {1},
  number  = {5},
  pages   = {AIdbp2300040},
  doi     = {10.1056/AIdbp2300040},
  note    = {doi:10.1056/AIdbp2300040}
}

@article{Jiang2025,
  author  = {Jiang, Y. and Black, K. C. and Geng, G. and Park, D. and Ng, A. Y. and Chen, J. H.},
  title   = {{MedAgentBench}: a virtual {EHR} environment to benchmark medical {LLM} agents},
  journal = {NEJM AI},
  year    = {2025},
  volume  = {2},
  number  = {2},
  pages   = {AIdbp2500144},
  doi     = {10.1056/AIdbp2500144},
  note    = {doi:10.1056/AIdbp2500144}
}

@article{Enikeev2026,
  author  = {Enikeev, R. and Moldovan, M. and Chu, M. and Amalraj, A. and Koli, P. P. and Syed Abdul, S. and Sivaraj, H. and Iqbal, U. and Toh, C. K.},
  title   = {Privacy-preserving large language model deployment for oncology registry abstraction: structure-aware evaluation in a real-world clinical setting},
  journal = {medRxiv},
  year    = {2026},
  pages   = {2026.05.18.26353541},
  doi     = {10.64898/2026.05.18.26353541},
  note    = {doi:10.64898/2026.05.18.26353541}
}

@incollection{Alwakeel2026,
  author    = {Alwakeel, M. and others},
  title     = {Evaluating large language models for automated clinical abstraction in pulmonary embolism registries: performance across model sizes, versions, and parameters},
  booktitle = {Applications of Medical Artificial Intelligence (AMAI 2025)},
  editor    = {Wu, S. and Shabestari, B. and Xing, L.},
  series    = {Lecture Notes in Computer Science},
  volume    = {16206},
  publisher = {Springer},
  address   = {Cham},
  year      = {2026},
  doi       = {10.1007/978-3-032-09569-5\_21},
  note    = {doi:10.1007/978-3-032-09569-5\_21}
}

@article{vanderLoo2026,
  author  = {van der Loo, W. and van der Valk, V. and van den Broek, T. and Atsma, D. and Staring, M. and Scherptong, R.},
  title   = {Large language models for structured cardiovascular data extraction: a foundation for scalable research and clinical applications},
  journal = {Eur Heart J Digit Health},
  year    = {2026},
  volume  = {7},
  number  = {2},
  pages   = {ztaf127},
  doi     = {10.1093/ehjdh/ztaf127},
  note    = {doi:10.1093/ehjdh/ztaf127}
}

@article{Brindis2001,
  author  = {Brindis, R. G. and Fitzgerald, S. and Anderson, H. V. and others},
  title   = {The American College of Cardiology--National Cardiovascular Data Registry ({ACC-NCDR}): building a national clinical data repository},
  journal = {J Am Coll Cardiol},
  year    = {2001},
  volume  = {37},
  number  = {8},
  pages   = {2240--2245},
  doi     = {10.1016/S0735-1097(01)01372-9},
  note    = {doi:10.1016/S0735-1097(01)01372-9}
}

@article{Chan2011,
  author  = {Chan, P. S. and Patel, M. R. and Klein, L. W. and others},
  title   = {Appropriateness of percutaneous coronary intervention},
  journal = {JAMA},
  year    = {2011},
  volume  = {306},
  number  = {1},
  pages   = {53--61},
  doi     = {10.1001/jama.2011.916},
  note    = {doi:10.1001/jama.2011.916}
}

@article{Hager2024,
  author  = {Hager, P. and Jungmann, F. and Holland, R. and others},
  title   = {Evaluation and mitigation of the limitations of large language models in clinical decision-making},
  journal = {Nat Med},
  year    = {2024},
  volume  = {30},
  number  = {9},
  pages   = {2613--2622},
  doi     = {10.1038/s41591-024-03097-1},
  note    = {doi:10.1038/s41591-024-03097-1}
}

@misc{OpenAI2025,
  author = {{OpenAI}},
  title  = {{HealthBench}: evaluating large language models across real-world healthcare conversations},
  year   = {2025},
  url    = {https://openai.com/index/healthbench/}
}

@article{Lee2025,
  author  = {Lee, D. and Vaid, A. and Menon, K. M. and Freeman, R. and Matteson, D. S. and Marin, M. L. and Nadkarni, G. N.},
  title   = {Using large language models to automate data extraction from surgical pathology reports: retrospective cohort study},
  journal = {JMIR Form Res},
  year    = {2025},
  volume  = {9},
  pages   = {e64544},
  doi     = {10.2196/64544},
  note    = {doi:10.2196/64544}
}

@misc{Kalai2025,
  author        = {Kalai, A. T. and Nachum, O. and Vempala, S. S. and Zhang, E.},
  title         = {Why language models hallucinate},
  year          = {2025},
  eprint        = {2509.04664},
  archivePrefix = {arXiv},
  primaryClass  = {cs.CL}
}

@article{Chung2026,
  author  = {Chung, P. and Swaminathan, A. and Goodell, A. J. and Kim, Y. and Momsen Reincke, S. and Han, L. and others},
  title   = {Verifying facts in patient care documents generated by large language models using electronic health records},
  journal = {NEJM AI},
  year    = {2026},
  volume  = {3},
  number  = {1},
  pages   = {AIdbp2500418}
}

@article{Fabacher2025,
  author  = {Fabacher, T. and Sauleau, E. A. and Arcay, E. and others},
  title   = {Efficient extraction of medication information from clinical notes: an evaluation in 2 languages},
  journal = {J Am Med Inform Assoc},
  year    = {2025},
  volume  = {32},
  number  = {12},
  pages   = {1855--1864},
  doi     = {10.1093/jamia/ocaf113},
  note    = {doi:10.1093/jamia/ocaf113}
}

@article{Sanghera2025,
  author  = {Sanghera, R. and Thirunavukarasu, A. J. and El Khoury, M. and others},
  title   = {High-performance automated abstract screening with large language model ensembles},
  journal = {J Am Med Inform Assoc},
  year    = {2025},
  volume  = {32},
  number  = {5},
  pages   = {893--904},
  doi     = {10.1093/jamia/ocaf050},
  note    = {doi:10.1093/jamia/ocaf050}
}

@article{OMalley2005,
  author  = {O'Malley, K. J. and Cook, K. F. and Price, M. D. and Wildes, K. R. and Hurdle, J. F. and Ashton, C. M.},
  title   = {Measuring diagnoses: {ICD} code accuracy},
  journal = {Health Serv Res},
  year    = {2005},
  volume  = {40},
  number  = {5 Pt 2},
  pages   = {1620--1639}
}

@article{Reamaroon2019,
  author  = {Reamaroon, N. and Sjoding, M. W. and Lin, K. and Iwashyna, T. J. and Najarian, K.},
  title   = {Accounting for label uncertainty in machine learning for detection of acute respiratory distress syndrome},
  journal = {IEEE J Biomed Health Inform},
  year    = {2019},
  volume  = {23},
  number  = {1},
  pages   = {407--415},
  doi     = {10.1109/JBHI.2018.2810820},
  note    = {doi:10.1109/JBHI.2018.2810820}
}

@article{Farquhar2024,
  author  = {Farquhar, S. and Kossen, J. and Kuhn, L. and Gal, Y.},
  title   = {Detecting hallucinations in large language models using semantic entropy},
  journal = {Nature},
  year    = {2024},
  volume  = {630},
  number  = {8017},
  pages   = {625--630}
}
\normalsize
\clearpage
\section*{Supplementary Appendix}
\noindent\textit{An ambiguity taxonomy for evaluating large language model performance on clinical registry abstraction: a multi-site prospective 
study}

\subsection*{Supplementary Methods}
\textbf{S1. Registry Background.} The pilot phase used the ACC NCDR Electrophysiology Device Implant registry (EPDI; formerly the Implantable 
Cardioverter-Defibrillator registry). The validation phase used the ACC NCDR Cardiac Catheterization and Percutaneous Coronary Intervention registry 
(CathPCI v5.7.1), which captures percutaneous coronary intervention data across more than 1,600 US hospitals. In the EPDI pilot, questions were 
grouped by registry section: Demographics; Episode of Care; History and Risk Factors; Diagnostic Studies; Labs; Procedure Information; Device 
Implant/Explant; Lead Assessment; Intra/Post-Procedure Events; and Discharge. This coarser scheme was suited to the pilot's document-routing 
objective.\par

\textbf{S2. Ambiguity Taxonomy: Full Category Definitions.} In the validation study, each registry question was assigned to one of six categories, 
ordered from lowest to highest reasoning demand. Medication/Event Flag (n = 62) comprised questions asking whether a specific medication was given or 
a specific peri-procedural event occurred, typically verifiable directly from structured documentation. Binary Clinical Presence (n = 41) comprised 
yes/no clinical conditions or comorbidities defined by ACC/NCDR coding criteria. Administrative/Transcription (n = 18) comprised demographic or 
identification data transcribed directly from the record without clinical interpretation. Quantitative Laboratory/Physiologic (n = 10) comprised 
numeric laboratory or physiologic measurements, occasionally requiring limited disambiguation when multiple sources exist. Clinical Interpretation (n 
= 22) required synthesis of multiple data sources and clinical judgment, such that reasonable disagreement between trained abstractors could occur. 
Event Timing (n = 4) required determining the correct clinical date or timestamp by identifying the moment from documentation that may conflict 
across source types.\par

\textbf{S3. Prompt Structure.} Each prompt consisted of three components: a general instruction simulating a clinical abstractor role, the verbatim 
ACC NCDR question definition (including value sets where specified), and a structured output format. Across every record-question pair, the model was 
given a single standing instruction, to answer when confident and to decline when not. For questions with defined value sets, the model was required 
to select from the allowed options or explicitly abstain; free-text substitutions were not permitted.\par

\textbf{S4. Software.} Analyses used Python 3.10 (NumPy 1.24, Pandas 2.0, SciPy 1.10).\par

\subsection*{Supplementary Material}
\vspace{0.5em}\noindent\textbf{Supplementary Table S1. Pilot Study: Per-Patient Detail} Pilot data (Institution A, EPDI registry) disaggregated by 
patient, showing how candidate-source counts and answerable rates varied across the three patient records. The cross-patient variance illustrates why 
source-data ambiguity must be controlled before measuring inherent task ambiguity: the same question can have very different document footprints 
across patients.\par\vspace{0.4em}
{\small
\begin{longtable}{c p{3.6cm} c c c}
\toprule
\textbf{Patient} & \textbf{Registry Section} & \textbf{N Questions} & \textbf{Total Answers (Mean $\pm$ SD)} & \textbf{Answerable \% (Mean $\pm$ SD)} 
\\
\midrule\endfirsthead
\toprule \textbf{Patient} & \textbf{Registry Section} & \textbf{N Questions} & \textbf{Total Answers (Mean $\pm$ SD)} & \textbf{Answerable \% (Mean 
$\pm$ SD)} \\ \midrule\endhead
Pat 1 & A. Demographics & 9 & 3.0 $\pm$ 0.0 & 44.4\% $\pm$ 23.6\% \\
Pat 1 & B. Episode of Care & 9 & 3.7 $\pm$ 1.6 & 36.7\% $\pm$ 33.2\% \\
Pat 1 & C. History and Risk Factors & 66 & 157.2 $\pm$ 22.1 & 54.4\% $\pm$ 14.8\% \\
Pat 1 & D. Diagnostic Studies & 13 & 57.0 $\pm$ 79.0 & 67.4\% $\pm$ 12.9\% \\
Pat 1 & E. Labs & 3 & 18.7 $\pm$ 0.6 & 81.8\% $\pm$ 18.1\% \\
Pat 1 & F. Procedure Information & 8 & 6.0 $\pm$ 2.5 & 49.3\% $\pm$ 15.4\% \\
Pat 1 & G. Device Implant/Explant & 23 & 7.8 $\pm$ 0.4 & 60.6\% $\pm$ 22.8\% \\
Pat 1 & H. Lead Assessment & 11 & 7.6 $\pm$ 0.5 & 54.2\% $\pm$ 23.4\% \\
Pat 1 & I. Intra/Post-Procedure Events & 15 & 14.5 $\pm$ 0.6 & 52.9\% $\pm$ 12.7\% \\
Pat 1 & J. Discharge & 10 & 8.3 $\pm$ 7.1 & 64.7\% $\pm$ 14.3\% \\
Pat 2 & A. Demographics & 9 & 33.7 $\pm$ 0.7 & 54.6\% $\pm$ 18.2\% \\
Pat 2 & B. Episode of Care & 9 & 34.7 $\pm$ 1.6 & 42.5\% $\pm$ 28.2\% \\
Pat 2 & C. History and Risk Factors & 66 & 85.0 $\pm$ 12.9 & 38.9\% $\pm$ 11.0\% \\
Pat 2 & D. Diagnostic Studies & 13 & 65.1 $\pm$ 32.0 & 77.2\% $\pm$ 20.8\% \\
Pat 2 & E. Labs & 3 & 33.7 $\pm$ 0.6 & 92.9\% $\pm$ 12.2\% \\
Pat 2 & F. Procedure Information & 8 & 50.8 $\pm$ 24.1 & 50.4\% $\pm$ 21.9\% \\
Pat 2 & G. Device Implant/Explant & 23 & 67.3 $\pm$ 9.5 & 63.3\% $\pm$ 20.0\% \\
Pat 2 & H. Lead Assessment & 11 & 62.8 $\pm$ 12.5 & 62.9\% $\pm$ 15.7\% \\
Pat 2 & I. Intra/Post-Procedure Events & 15 & 54.6 $\pm$ 0.8 & 53.6\% $\pm$ 14.1\% \\
Pat 2 & J. Discharge & 10 & 20.9 $\pm$ 23.6 & 81.4\% $\pm$ 16.9\% \\
Pat 3 & A. Demographics & 9 & 7.0 $\pm$ 0.0 & 52.4\% $\pm$ 16.0\% \\
Pat 3 & B. Episode of Care & 9 & 7.1 $\pm$ 1.1 & 41.7\% $\pm$ 27.0\% \\
Pat 3 & C. History and Risk Factors & 66 & 25.3 $\pm$ 3.2 & 33.2\% $\pm$ 8.5\% \\
Pat 3 & D. Diagnostic Studies & 13 & 23.2 $\pm$ 3.7 & 47.4\% $\pm$ 18.1\% \\
Pat 3 & E. Labs & 3 & 23.0 $\pm$ 0.0 & 62.3\% $\pm$ 13.3\% \\
Pat 3 & F. Procedure Information & 8 & 9.6 $\pm$ 3.9 & 51.2\% $\pm$ 15.7\% \\
Pat 3 & G. Device Implant/Explant & 23 & 12.4 $\pm$ 1.2 & 57.0\% $\pm$ 21.1\% \\
Pat 3 & H. Lead Assessment & 11 & 11.9 $\pm$ 1.5 & 56.0\% $\pm$ 15.7\% \\
Pat 3 & I. Intra/Post-Procedure Events & 15 & 18.0 $\pm$ 0.0 & 35.2\% $\pm$ 9.5\% \\
Pat 3 & J. Discharge & 10 & 8.7 $\pm$ 9.0 & 79.0\% $\pm$ 27.6\% \\
\bottomrule
\end{longtable}}

\vspace{0.5em}\noindent\textbf{Supplementary Table S2. Document-Context Targeting: Detailed Retrieval Specifications} For each documentation context 
summarized in Table 2, this table gives the FHIR resources, document types, and example questions used by the LLM during the validation 
study.\par\vspace{0.4em}
{\small
\begin{longtable}{p{2.6cm} p{4.2cm} c p{4.2cm}}
\toprule
\textbf{Document Context} & \textbf{FHIR Resources / Document Types Retrieved} & \textbf{\# Questions} & \textbf{Example Questions} \\
\midrule\endfirsthead
\toprule \textbf{Document Context} & \textbf{FHIR Resources / Document Types Retrieved} & \textbf{\# Questions} & \textbf{Example Questions} \\ 
\midrule\endhead
All Documents & Full patient record, no filters applied & 3 & Arrival Date/Time; Discharge Date/Time; Procedure Start Date/Time \\ \addlinespace
Patient Demographics \& H\&P & Patient resource; Encounter; History \& Physical; Consult notes; Discharge summaries & 16 & Patient Last/First Name; 
DOB; Sex; Race; Hispanic Origin; Health Insurance; ZIP Code \\ \addlinespace
Broad Clinical Notes & Encounter; Progress notes; Discharge summaries; H\&P; Consult; Procedure; Nursing; Transfer summaries & 111 & Admission 
Source; Prior MI; Hypertension; Diabetes; Dyslipidemia; NYHA Class; Prior PCI; Tobacco Use; Stress Test Type/Result \\ \addlinespace
Lab Results & Lab Observations; Diagnostic reports (CBC, metabolic panels, cath-lab labs) & 12 & Hemoglobin; Sodium; Creatinine (pre/post-procedure); 
Potassium; BUN; Calcium Score Assessed \\ \addlinespace
Medications & MedicationRequest; MedicationAdministration; MedicationStatement; Discharge summaries & 5 & Discharge Medication Code (RxNorm); 
Pre-procedure / Procedure Medication Administration; GDMT Maximum Dose \\ \addlinespace
Discharge Summary & Discharge summary documents only & 10 & CABG Status; PCI Date; Discharge Status/Location/Hospice; Comfort Care; Cardiac Rehab 
Referral \\ \addlinespace
Echocardiography \& Imaging & Echocardiography; cardiac ultrasound reports & 4 & Prior LVEF Assessed; Most Recent LVEF \%/Date; ECG Results \\ 
\addlinespace
\bottomrule
\end{longtable}}

\vspace{0.5em}\noindent\textbf{Supplementary Table S3. Complete Ranked Question-Level Performance (n $\geq$ 20)} Full 157-question performance 
distribution from the CathPCI validation study, ordered by ascending accuracy. PR Gap = Recall - Precision. Question identifiers are given as the 
registry element names.\par\vspace{0.4em}
{\footnotesize\setlength{\tabcolsep}{4pt}
\begin{longtable}{r >{\raggedright\arraybackslash}p{4.3cm} >{\raggedright\arraybackslash}p{2.2cm} c r r r r}
\toprule
\textbf{Rank} & \textbf{Question ID} & \textbf{Category} & \textbf{N} & \textbf{Accuracy} & \textbf{Precision} & \textbf{Recall} & \textbf{PR Gap} \\
\midrule\endfirsthead
\multicolumn{8}{l}{\footnotesize\textit{Supplementary Table S3 (continued)}}\\ \toprule \textbf{Rank} & \textbf{Question ID} & \textbf{Category} & 
\textbf{N} & \textbf{Accuracy} & \textbf{Precision} & \textbf{Recall} & \textbf{PR Gap} \\ \midrule\endhead
\midrule \multicolumn{8}{r}{\footnotesize\textit{continued on next page}}\\ \endfoot
\bottomrule\endlastfoot
1 & cath\_\allowbreak{}lab\_\allowbreak{}visit\_\allowbreak{}indication & Clinical Interpretation & 25 & 12.0\% & 12.0\% & 100.0\% & 88.0\% \\
2 & arrival\_\allowbreak{}date\_\allowbreak{}time & Event Timing & 25 & 20.0\% & 20.0\% & 100.0\% & 80.0\% \\
3 & adm\_\allowbreak{}l\_\allowbreak{}name & Administrative & 25 & 48.0\% & 50.0\% & 92.3\% & 42.3\% \\
4 & csha\_\allowbreak{}scale & Clinical Interpretation & 25 & 56.0\% & 56.0\% & 100.0\% & 44.0\% \\
5 & cv\_\allowbreak{}instability & Clinical Interpretation & 25 & 56.0\% & 56.0\% & 100.0\% & 44.0\% \\
6 & dc\_\allowbreak{}date\_\allowbreak{}time & Event Timing & 25 & 60.0\% & 60.0\% & 100.0\% & 40.0\% \\
7 & pci\_\allowbreak{}indication & Clinical Interpretation & 25 & 68.0\% & 68.0\% & 100.0\% & 32.0\% \\
8 & hx\_\allowbreak{}mi & Binary Clinical Presence & 25 & 68.0\% & 68.0\% & 100.0\% & 32.0\% \\
9 & pre\_\allowbreak{}proc\_\allowbreak{}lvef\_\allowbreak{}assessed & Clinical Interpretation & 25 & 68.0\% & 68.0\% & 100.0\% & 32.0\% \\
10 & cabg\_\allowbreak{}planned\_\allowbreak{}dc & Clinical Interpretation & 24 & 70.8\% & 94.4\% & 73.9\% & 20.5\% \\
11 & multi\_\allowbreak{}vessel\_\allowbreak{}dz & Clinical Interpretation & 25 & 72.0\% & 72.0\% & 100.0\% & 28.0\% \\
12 & pci\_\allowbreak{}status & Clinical Interpretation & 25 & 72.0\% & 72.0\% & 100.0\% & 28.0\% \\
13 & ec\_\allowbreak{}assess\_\allowbreak{}method & Clinical Interpretation & 25 & 72.0\% & 72.0\% & 100.0\% & 28.0\% \\
14 & weight & Quantitative Lab/Physiologic & 25 & 72.0\% & 78.3\% & 90.0\% & 11.7\% \\
15 & d\_\allowbreak{}cath\_\allowbreak{}l\_\allowbreak{}name & Administrative & 24 & 75.0\% & 75.0\% & 100.0\% & 25.0\% \\
16 & dyslipidemia & Binary Clinical Presence & 25 & 76.0\% & 76.0\% & 100.0\% & 24.0\% \\
17 & dc\_\allowbreak{}med.\allowbreak{}41549009.\allowbreak{}dc\_\allowbreak{}med\_\allowbreak{}admin & Medication/Event Flag & 23 & 78.3\% & 78.3\% 
& 100.0\% & 21.7\% \\
18 & pre\_\allowbreak{}proc\_\allowbreak{}timi & Clinical Interpretation & 24 & 79.2\% & 86.4\% & 90.5\% & 4.1\% \\
19 & prior\_\allowbreak{}dx\_\allowbreak{}angio\_\allowbreak{}proc & Binary Clinical Presence & 25 & 80.0\% & 80.0\% & 100.0\% & 20.0\% \\
20 & post\_\allowbreak{}proc\_\allowbreak{}creat & Quantitative Lab/Physiologic & 20 & 80.0\% & 80.0\% & 100.0\% & 20.0\% \\
21 & stenosis\_\allowbreak{}prior\_\allowbreak{}treat & Clinical Interpretation & 25 & 80.0\% & 80.0\% & 100.0\% & 20.0\% \\
22 & stress\_\allowbreak{}test\_\allowbreak{}result & Clinical Interpretation & 20 & 80.0\% & 84.2\% & 94.1\% & 9.9\% \\
23 & procedure\_\allowbreak{}end\_\allowbreak{}date\_\allowbreak{}time & Event Timing & 25 & 80.0\% & 80.0\% & 100.0\% & 20.0\% \\
24 & stress\_\allowbreak{}test\_\allowbreak{}type & Clinical Interpretation & 20 & 80.0\% & 84.2\% & 94.1\% & 9.9\% \\
25 & dc\_\allowbreak{}med.\allowbreak{}33252009.\allowbreak{}dc\_\allowbreak{}med\_\allowbreak{}admin & Medication/Event Flag & 23 & 82.6\% & 82.6\% 
& 100.0\% & 17.4\% \\
26 & dc\_\allowbreak{}med\_\allowbreak{}reconciled & Clinical Interpretation & 23 & 82.6\% & 86.4\% & 95.0\% & 8.6\% \\
27 & dc\_\allowbreak{}med\_\allowbreak{}recon\_\allowbreak{}completed & Clinical Interpretation & 23 & 82.6\% & 82.6\% & 100.0\% & 17.4\% \\
28 & dc\_\allowbreak{}card\_\allowbreak{}rehab & Clinical Interpretation & 24 & 83.3\% & 83.3\% & 100.0\% & 16.7\% \\
29 & ipp\_\allowbreak{}event.\allowbreak{}385494008.\allowbreak{}post\_\allowbreak{}proc\_\allowbreak{}occurred & Medication/Event Flag & 25 & 84.0\% 
& 84.0\% & 100.0\% & 16.0\% \\
30 & pre\_\allowbreak{}proc\_\allowbreak{}med.\allowbreak{}1191.\allowbreak{}pre\_\allowbreak{}proc\_\allowbreak{}med\_\allowbreak{}admin & 
Medication/Event Flag & 25 & 84.0\% & 84.0\% & 100.0\% & 16.0\% \\
31 & hgb & Quantitative Lab/Physiologic & 21 & 85.7\% & 90.0\% & 94.7\% & 4.7\% \\
32 & dc\_\allowbreak{}med.\allowbreak{}372913009.\allowbreak{}dc\_\allowbreak{}med\_\allowbreak{}admin & Medication/Event Flag & 23 & 87.0\% & 87.0\% 
& 100.0\% & 13.0\% \\
33 & ipp\_\allowbreak{}event.\allowbreak{}1000142371.\allowbreak{}post\_\allowbreak{}proc\_\allowbreak{}occurred & Medication/Event Flag & 25 & 
88.0\% & 88.0\% & 100.0\% & 12.0\% \\
34 & ipp\_\allowbreak{}event.\allowbreak{}1000142419.\allowbreak{}post\_\allowbreak{}proc\_\allowbreak{}occurred & Medication/Event Flag & 25 & 
88.0\% & 88.0\% & 100.0\% & 12.0\% \\
35 & hx\_\allowbreak{}cvd & Binary Clinical Presence & 25 & 88.0\% & 88.0\% & 100.0\% & 12.0\% \\
36 & prior\_\allowbreak{}pad & Binary Clinical Presence & 25 & 88.0\% & 88.0\% & 100.0\% & 12.0\% \\
37 & pre\_\allowbreak{}proc\_\allowbreak{}med.\allowbreak{}33252009.\allowbreak{}pre\_\allowbreak{}proc\_\allowbreak{}med\_\allowbreak{}admin & 
Medication/Event Flag & 25 & 88.0\% & 88.0\% & 100.0\% & 12.0\% \\
38 & pcil\_\allowbreak{}name & Administrative & 25 & 88.0\% & 88.0\% & 100.0\% & 12.0\% \\
39 & cardiac\_\allowbreak{}cta & Binary Clinical Presence & 25 & 88.0\% & 88.0\% & 100.0\% & 12.0\% \\
40 & ipp\_\allowbreak{}event.\allowbreak{}1000142440.\allowbreak{}post\_\allowbreak{}proc\_\allowbreak{}occurred & Medication/Event Flag & 25 & 
88.0\% & 88.0\% & 100.0\% & 12.0\% \\
41 & proc\_\allowbreak{}systolic\_\allowbreak{}bp & Quantitative Lab/Physiologic & 25 & 88.0\% & 91.7\% & 95.6\% & 4.0\% \\
42 & procedure\_\allowbreak{}start\_\allowbreak{}date\_\allowbreak{}time & Event Timing & 25 & 88.0\% & 88.0\% & 100.0\% & 12.0\% \\
43 & family\_\allowbreak{}hx\_\allowbreak{}cad & Binary Clinical Presence & 25 & 88.0\% & 91.7\% & 95.6\% & 4.0\% \\
44 & pre\_\allowbreak{}proc\_\allowbreak{}creat & Quantitative Lab/Physiologic & 22 & 90.9\% & 90.9\% & 100.0\% & 9.1\% \\
45 & anti\_\allowbreak{}arrhy\_\allowbreak{}therapy & Clinical Interpretation & 22 & 90.9\% & 95.2\% & 95.2\% & 0.0\% \\
46 & dc\_\allowbreak{}med.\allowbreak{}1116632.\allowbreak{}dc\_\allowbreak{}med\_\allowbreak{}admin & Medication/Event Flag & 23 & 91.3\% & 91.3\% & 
100.0\% & 8.7\% \\
47 & pre\_\allowbreak{}proc\_\allowbreak{}med.\allowbreak{}96302009.\allowbreak{}pre\_\allowbreak{}proc\_\allowbreak{}med\_\allowbreak{}admin & 
Medication/Event Flag & 25 & 92.0\% & 92.0\% & 100.0\% & 8.0\% \\
48 & calcium\_\allowbreak{}score\_\allowbreak{}assessed & Binary Clinical Presence & 25 & 92.0\% & 92.0\% & 100.0\% & 8.0\% \\
49 & height & Quantitative Lab/Physiologic & 25 & 92.0\% & 100.0\% & 92.0\% & 8.0\% \\
50 & access\_\allowbreak{}site & Binary Clinical Presence & 25 & 92.0\% & 92.0\% & 100.0\% & 8.0\% \\
51 & pre\_\allowbreak{}proc\_\allowbreak{}med.\allowbreak{}372913009.\allowbreak{}pre\_\allowbreak{}proc\_\allowbreak{}med\_\allowbreak{}admin & 
Medication/Event Flag & 25 & 92.0\% & 92.0\% & 100.0\% & 8.0\% \\
52 & pre\_\allowbreak{}proc\_\allowbreak{}med.\allowbreak{}41549009.\allowbreak{}pre\_\allowbreak{}proc\_\allowbreak{}med\_\allowbreak{}admin & 
Medication/Event Flag & 25 & 92.0\% & 92.0\% & 100.0\% & 8.0\% \\
53 & ipp\_\allowbreak{}event.\allowbreak{}22298006.\allowbreak{}post\_\allowbreak{}proc\_\allowbreak{}occurred & Medication/Event Flag & 25 & 92.0\% 
& 92.0\% & 100.0\% & 8.0\% \\
54 & ipp\_\allowbreak{}event.\allowbreak{}84114007.\allowbreak{}post\_\allowbreak{}proc\_\allowbreak{}occurred & Medication/Event Flag & 25 & 92.0\% 
& 92.0\% & 100.0\% & 8.0\% \\
55 & concom\_\allowbreak{}proc & Binary Clinical Presence & 25 & 92.0\% & 92.0\% & 100.0\% & 8.0\% \\
56 & pci\_\allowbreak{}decision & Clinical Interpretation & 25 & 92.0\% & 92.0\% & 100.0\% & 8.0\% \\
57 & hx\_\allowbreak{}hf & Binary Clinical Presence & 25 & 92.0\% & 92.0\% & 100.0\% & 8.0\% \\
58 & dc\_\allowbreak{}med.\allowbreak{}96302009.\allowbreak{}dc\_\allowbreak{}med\_\allowbreak{}admin & Medication/Event Flag & 23 & 95.6\% & 95.6\% 
& 100.0\% & 4.3\% \\
59 & dc\_\allowbreak{}med.\allowbreak{}1191.\allowbreak{}dc\_\allowbreak{}med\_\allowbreak{}admin & Medication/Event Flag & 23 & 95.6\% & 95.6\% & 
100.0\% & 4.3\% \\
60 & dc\_\allowbreak{}med.\allowbreak{}100014161.\allowbreak{}dc\_\allowbreak{}med\_\allowbreak{}admin & Medication/Event Flag & 23 & 95.6\% & 95.6\% 
& 100.0\% & 4.3\% \\
61 & dc\_\allowbreak{}med.\allowbreak{}1546356.\allowbreak{}dc\_\allowbreak{}med\_\allowbreak{}admin & Medication/Event Flag & 23 & 95.6\% & 95.6\% & 
100.0\% & 4.3\% \\
62 & dc\_\allowbreak{}med.\allowbreak{}1659152.\allowbreak{}dc\_\allowbreak{}med\_\allowbreak{}admin & Medication/Event Flag & 23 & 95.6\% & 95.6\% & 
100.0\% & 4.3\% \\
63 & dc\_\allowbreak{}med.\allowbreak{}613391.\allowbreak{}dc\_\allowbreak{}med\_\allowbreak{}admin & Medication/Event Flag & 23 & 95.6\% & 95.6\% & 
100.0\% & 4.3\% \\
64 & dc\_\allowbreak{}med.\allowbreak{}1665684.\allowbreak{}dc\_\allowbreak{}med\_\allowbreak{}admin & Medication/Event Flag & 23 & 95.6\% & 95.6\% & 
100.0\% & 4.3\% \\
65 & dc\_\allowbreak{}hospice & Binary Clinical Presence & 24 & 95.8\% & 95.8\% & 100.0\% & 4.2\% \\
66 & proc\_\allowbreak{}med.\allowbreak{}1116632.\allowbreak{}proc\_\allowbreak{}med\_\allowbreak{}admin & Medication/Event Flag & 25 & 96.0\% & 
96.0\% & 100.0\% & 4.0\% \\
67 & proc\_\allowbreak{}med.\allowbreak{}1000142427.\allowbreak{}proc\_\allowbreak{}med\_\allowbreak{}admin & Medication/Event Flag & 25 & 96.0\% & 
100.0\% & 96.0\% & 4.0\% \\
68 & ecg\_\allowbreak{}results & Clinical Interpretation & 25 & 96.0\% & 100.0\% & 96.0\% & 4.0\% \\
69 & contrast\_\allowbreak{}vol & Quantitative Lab/Physiologic & 25 & 96.0\% & 100.0\% & 96.0\% & 4.0\% \\
70 & ca\_\allowbreak{}in\_\allowbreak{}hosp & Binary Clinical Presence & 25 & 96.0\% & 96.0\% & 100.0\% & 4.0\% \\
71 & current\_\allowbreak{}dialysis & Binary Clinical Presence & 25 & 96.0\% & 96.0\% & 100.0\% & 4.0\% \\
72 & crossover & Binary Clinical Presence & 25 & 96.0\% & 96.0\% & 100.0\% & 4.0\% \\
73 & dc\_\allowbreak{}comfort & Binary Clinical Presence & 25 & 96.0\% & 96.0\% & 100.0\% & 4.0\% \\
74 & ipp\_\allowbreak{}event.\allowbreak{}410429000.\allowbreak{}post\_\allowbreak{}proc\_\allowbreak{}occurred & Medication/Event Flag & 25 & 96.0\% 
& 96.0\% & 100.0\% & 4.0\% \\
75 & left\_\allowbreak{}heart\_\allowbreak{}cath & Binary Clinical Presence & 25 & 96.0\% & 96.0\% & 100.0\% & 4.0\% \\
76 & prior\_\allowbreak{}pci & Binary Clinical Presence & 25 & 96.0\% & 96.0\% & 100.0\% & 4.0\% \\
77 & ipp\_\allowbreak{}event.\allowbreak{}100014076.\allowbreak{}post\_\allowbreak{}proc\_\allowbreak{}occurred & Medication/Event Flag & 25 & 96.0\% 
& 96.0\% & 100.0\% & 4.0\% \\
78 & hx\_\allowbreak{}chronic\_\allowbreak{}lung\_\allowbreak{}disease & Binary Clinical Presence & 25 & 96.0\% & 96.0\% & 100.0\% & 4.0\% \\
79 & venous\_\allowbreak{}access & Binary Clinical Presence & 25 & 96.0\% & 96.0\% & 100.0\% & 4.0\% \\
80 & v\_\allowbreak{}support & Binary Clinical Presence & 25 & 96.0\% & 96.0\% & 100.0\% & 4.0\% \\
81 & zip\_\allowbreak{}code & Administrative & 25 & 96.0\% & 96.0\% & 100.0\% & 4.0\% \\
82 & race\_\allowbreak{}asian & Administrative & 25 & 96.0\% & 100.0\% & 96.0\% & 4.0\% \\
83 & proc\_\allowbreak{}med.\allowbreak{}32968.\allowbreak{}proc\_\allowbreak{}med\_\allowbreak{}admin & Medication/Event Flag & 25 & 96.0\% & 96.0\% 
& 100.0\% & 4.0\% \\
84 & proc\_\allowbreak{}med.\allowbreak{}96382006.\allowbreak{}proc\_\allowbreak{}med\_\allowbreak{}admin & Medication/Event Flag & 25 & 96.0\% & 
100.0\% & 96.0\% & 4.0\% \\
85 & race\_\allowbreak{}white & Administrative & 25 & 96.0\% & 100.0\% & 96.0\% & 4.0\% \\
86 & segment\_\allowbreak{}id & Clinical Interpretation & 25 & 96.0\% & 96.0\% & 100.0\% & 4.0\% \\
87 & tobacco\_\allowbreak{}use & Binary Clinical Presence & 25 & 96.0\% & 100.0\% & 96.0\% & 4.0\% \\
88 & race\_\allowbreak{}nat\_\allowbreak{}haw & Administrative & 25 & 96.0\% & 100.0\% & 96.0\% & 4.0\% \\
89 & proc\_\allowbreak{}med.\allowbreak{}400610005.\allowbreak{}proc\_\allowbreak{}med\_\allowbreak{}admin & Medication/Event Flag & 25 & 96.0\% & 
100.0\% & 96.0\% & 4.0\% \\
90 & ipp\_\allowbreak{}event.\allowbreak{}89138009.\allowbreak{}post\_\allowbreak{}proc\_\allowbreak{}occurred & Medication/Event Flag & 25 & 96.0\% 
& 96.0\% & 100.0\% & 4.0\% \\
91 & fluoro\_\allowbreak{}dose\_\allowbreak{}dap & Quantitative Lab/Physiologic & 25 & 96.0\% & 96.0\% & 100.0\% & 4.0\% \\
92 & graft\_\allowbreak{}stenosis & Binary Clinical Presence & 25 & 96.0\% & 96.0\% & 100.0\% & 4.0\% \\
93 & health\_\allowbreak{}ins & Administrative & 25 & 96.0\% & 96.0\% & 100.0\% & 4.0\% \\
94 & diag\_\allowbreak{}cor\_\allowbreak{}angio & Binary Clinical Presence & 25 & 96.0\% & 96.0\% & 100.0\% & 4.0\% \\
95 & first\_\allowbreak{}name & Administrative & 25 & 96.0\% & 96.0\% & 100.0\% & 4.0\% \\
96 & dominance & Clinical Interpretation & 25 & 96.0\% & 96.0\% & 100.0\% & 4.0\% \\
97 & fluoro\_\allowbreak{}dose\_\allowbreak{}kerm & Quantitative Lab/Physiologic & 25 & 100.0\% & 100.0\% & 100.0\% & 0.0\% \\
98 & pci\_\allowbreak{}proc & Binary Clinical Presence & 25 & 100.0\% & 100.0\% & 100.0\% & 0.0\% \\
99 & last\_\allowbreak{}name & Administrative & 25 & 100.0\% & 100.0\% & 100.0\% & 0.0\% \\
100 & nv\_\allowbreak{}stenosis & Binary Clinical Presence & 25 & 100.0\% & 100.0\% & 100.0\% & 0.0\% \\
101 & ipp\_\allowbreak{}event.\allowbreak{}422504002.\allowbreak{}post\_\allowbreak{}proc\_\allowbreak{}occurred & Medication/Event Flag & 25 & 
100.0\% & 100.0\% & 100.0\% & 0.0\% \\
102 & ipp\_\allowbreak{}event.\allowbreak{}417941003.\allowbreak{}post\_\allowbreak{}proc\_\allowbreak{}occurred & Medication/Event Flag & 25 & 
100.0\% & 100.0\% & 100.0\% & 0.0\% \\
103 & ipp\_\allowbreak{}event.\allowbreak{}74474003.\allowbreak{}post\_\allowbreak{}proc\_\allowbreak{}occurred & Medication/Event Flag & 25 & 
100.0\% & 100.0\% & 100.0\% & 0.0\% \\
104 & mid\_\allowbreak{}name & Administrative & 23 & 100.0\% & 100.0\% & 100.0\% & 0.0\% \\
105 & hypertension & Binary Clinical Presence & 25 & 100.0\% & 100.0\% & 100.0\% & 0.0\% \\
106 & hips & Binary Clinical Presence & 24 & 100.0\% & 100.0\% & 100.0\% & 0.0\% \\
107 & hisp\_\allowbreak{}orig & Administrative & 25 & 100.0\% & 100.0\% & 100.0\% & 0.0\% \\
108 & hosp\_\allowbreak{}intervention & Binary Clinical Presence & 25 & 100.0\% & 100.0\% & 100.0\% & 0.0\% \\
109 & ipp\_\allowbreak{}event.\allowbreak{}230713003.\allowbreak{}post\_\allowbreak{}proc\_\allowbreak{}occurred & Medication/Event Flag & 25 & 
100.0\% & 100.0\% & 100.0\% & 0.0\% \\
110 & ipp\_\allowbreak{}event.\allowbreak{}230706003.\allowbreak{}post\_\allowbreak{}proc\_\allowbreak{}occurred & Medication/Event Flag & 25 & 
100.0\% & 100.0\% & 100.0\% & 0.0\% \\
111 & dissection\_\allowbreak{}seg & Binary Clinical Presence & 25 & 100.0\% & 100.0\% & 100.0\% & 0.0\% \\
112 & fluoro\_\allowbreak{}time & Quantitative Lab/Physiologic & 25 & 100.0\% & 100.0\% & 100.0\% & 0.0\% \\
113 & ca\_\allowbreak{}out\_\allowbreak{}hospital & Binary Clinical Presence & 25 & 100.0\% & 100.0\% & 100.0\% & 0.0\% \\
114 & cp\_\allowbreak{}sx\_\allowbreak{}assess & Binary Clinical Presence & 25 & 100.0\% & 100.0\% & 100.0\% & 0.0\% \\
115 & ca\_\allowbreak{}transfer\_\allowbreak{}fac & Binary Clinical Presence & 25 & 100.0\% & 100.0\% & 100.0\% & 0.0\% \\
116 & enrolled\_\allowbreak{}study & Administrative & 25 & 100.0\% & 100.0\% & 100.0\% & 0.0\% \\
117 & dob & Administrative & 25 & 100.0\% & 100.0\% & 100.0\% & 0.0\% \\
118 & dc\_\allowbreak{}status & Binary Clinical Presence & 25 & 100.0\% & 100.0\% & 100.0\% & 0.0\% \\
119 & device\_\allowbreak{}deployed & Binary Clinical Presence & 25 & 100.0\% & 100.0\% & 100.0\% & 0.0\% \\
120 & diabetes & Binary Clinical Presence & 25 & 100.0\% & 100.0\% & 100.0\% & 0.0\% \\
121 & dc\_\allowbreak{}med.\allowbreak{}32968.\allowbreak{}dc\_\allowbreak{}med\_\allowbreak{}admin & Medication/Event Flag & 23 & 100.0\% & 100.0\% 
& 100.0\% & 0.0\% \\
122 & dc\_\allowbreak{}med.\allowbreak{}1364430.\allowbreak{}dc\_\allowbreak{}med\_\allowbreak{}admin & Medication/Event Flag & 23 & 100.0\% & 
100.0\% & 100.0\% & 0.0\% \\
123 & dc\_\allowbreak{}med.\allowbreak{}1537034.\allowbreak{}dc\_\allowbreak{}med\_\allowbreak{}admin & Medication/Event Flag & 23 & 100.0\% & 
100.0\% & 100.0\% & 0.0\% \\
124 & dc\_\allowbreak{}med.\allowbreak{}1599538.\allowbreak{}dc\_\allowbreak{}med\_\allowbreak{}admin & Medication/Event Flag & 23 & 100.0\% & 
100.0\% & 100.0\% & 0.0\% \\
125 & dc\_\allowbreak{}location & Binary Clinical Presence & 24 & 100.0\% & 100.0\% & 100.0\% & 0.0\% \\
126 & dc\_\allowbreak{}med.\allowbreak{}10594.\allowbreak{}dc\_\allowbreak{}med\_\allowbreak{}admin & Medication/Event Flag & 23 & 100.0\% & 100.0\% 
& 100.0\% & 0.0\% \\
127 & dc\_\allowbreak{}med.\allowbreak{}1114195.\allowbreak{}dc\_\allowbreak{}med\_\allowbreak{}admin & Medication/Event Flag & 23 & 100.0\% & 
100.0\% & 100.0\% & 0.0\% \\
128 & dc\_\allowbreak{}med.\allowbreak{}11289.\allowbreak{}dc\_\allowbreak{}med\_\allowbreak{}admin & Medication/Event Flag & 23 & 100.0\% & 100.0\% 
& 100.0\% & 0.0\% \\
129 & proc\_\allowbreak{}med.\allowbreak{}1537034.\allowbreak{}proc\_\allowbreak{}med\_\allowbreak{}admin & Medication/Event Flag & 25 & 100.0\% & 
100.0\% & 100.0\% & 0.0\% \\
130 & proc\_\allowbreak{}med.\allowbreak{}15202.\allowbreak{}proc\_\allowbreak{}med\_\allowbreak{}admin & Medication/Event Flag & 25 & 100.0\% & 
100.0\% & 100.0\% & 0.0\% \\
131 & proc\_\allowbreak{}med.\allowbreak{}1364430.\allowbreak{}proc\_\allowbreak{}med\_\allowbreak{}admin & Medication/Event Flag & 25 & 100.0\% & 
100.0\% & 100.0\% & 0.0\% \\
132 & proc\_\allowbreak{}med.\allowbreak{}11289.\allowbreak{}proc\_\allowbreak{}med\_\allowbreak{}admin & Medication/Event Flag & 25 & 100.0\% & 
100.0\% & 100.0\% & 0.0\% \\
133 & proc\_\allowbreak{}med.\allowbreak{}1114195.\allowbreak{}proc\_\allowbreak{}med\_\allowbreak{}admin & Medication/Event Flag & 25 & 100.0\% & 
100.0\% & 100.0\% & 0.0\% \\
134 & prior\_\allowbreak{}cabg & Binary Clinical Presence & 25 & 100.0\% & 100.0\% & 100.0\% & 0.0\% \\
135 & pre\_\allowbreak{}proc\_\allowbreak{}med.\allowbreak{}48698004.\allowbreak{}pre\_\allowbreak{}proc\_\allowbreak{}med\_\allowbreak{}admin & 
Medication/Event Flag & 25 & 100.0\% & 100.0\% & 100.0\% & 0.0\% \\
136 & perf\_\allowbreak{}seg & Binary Clinical Presence & 25 & 100.0\% & 100.0\% & 100.0\% & 0.0\% \\
137 & post\_\allowbreak{}proc\_\allowbreak{}timi & Clinical Interpretation & 24 & 100.0\% & 100.0\% & 100.0\% & 0.0\% \\
138 & post\_\allowbreak{}transfusion & Binary Clinical Presence & 25 & 100.0\% & 100.0\% & 100.0\% & 0.0\% \\
139 & pre\_\allowbreak{}proc\_\allowbreak{}med.\allowbreak{}100014162.\allowbreak{}pre\_\allowbreak{}proc\_\allowbreak{}med\_\allowbreak{}admin & 
Medication/Event Flag & 25 & 100.0\% & 100.0\% & 100.0\% & 0.0\% \\
140 & pre\_\allowbreak{}proc\_\allowbreak{}med.\allowbreak{}100014161.\allowbreak{}pre\_\allowbreak{}proc\_\allowbreak{}med\_\allowbreak{}admin & 
Medication/Event Flag & 25 & 100.0\% & 100.0\% & 100.0\% & 0.0\% \\
141 & pre\_\allowbreak{}proc\_\allowbreak{}med.\allowbreak{}112000000694.\allowbreak{}pre\_\allowbreak{}proc\_\allowbreak{}med\_\allowbreak{}admin & 
Medication/Event Flag & 25 & 100.0\% & 100.0\% & 100.0\% & 0.0\% \\
142 & pre\_\allowbreak{}proc\_\allowbreak{}med.\allowbreak{}1656341.\allowbreak{}pre\_\allowbreak{}proc\_\allowbreak{}med\_\allowbreak{}admin & 
Medication/Event Flag & 25 & 100.0\% & 100.0\% & 100.0\% & 0.0\% \\
143 & pre\_\allowbreak{}proc\_\allowbreak{}med.\allowbreak{}31970009.\allowbreak{}pre\_\allowbreak{}proc\_\allowbreak{}med\_\allowbreak{}admin & 
Medication/Event Flag & 25 & 100.0\% & 100.0\% & 100.0\% & 0.0\% \\
144 & pre\_\allowbreak{}proc\_\allowbreak{}med.\allowbreak{}35829.\allowbreak{}pre\_\allowbreak{}proc\_\allowbreak{}med\_\allowbreak{}admin & 
Medication/Event Flag & 25 & 100.0\% & 100.0\% & 100.0\% & 0.0\% \\
145 & ipp\_\allowbreak{}event.\allowbreak{}95549001.\allowbreak{}post\_\allowbreak{}proc\_\allowbreak{}occurred & Medication/Event Flag & 25 & 
100.0\% & 100.0\% & 100.0\% & 0.0\% \\
146 & ipp\_\allowbreak{}event.\allowbreak{}35304003.\allowbreak{}post\_\allowbreak{}proc\_\allowbreak{}occurred & Medication/Event Flag & 25 & 
100.0\% & 100.0\% & 100.0\% & 0.0\% \\
147 & race\_\allowbreak{}black & Administrative & 25 & 100.0\% & 100.0\% & 100.0\% & 0.0\% \\
148 & proc\_\allowbreak{}med.\allowbreak{}1599538.\allowbreak{}proc\_\allowbreak{}med\_\allowbreak{}admin & Medication/Event Flag & 25 & 100.0\% & 
100.0\% & 100.0\% & 0.0\% \\
149 & proc\_\allowbreak{}med.\allowbreak{}321208.\allowbreak{}proc\_\allowbreak{}med\_\allowbreak{}admin & Medication/Event Flag & 25 & 100.0\% & 
100.0\% & 100.0\% & 0.0\% \\
150 & proc\_\allowbreak{}med.\allowbreak{}1656052.\allowbreak{}proc\_\allowbreak{}med\_\allowbreak{}admin & Medication/Event Flag & 25 & 100.0\% & 
100.0\% & 100.0\% & 0.0\% \\
151 & proc\_\allowbreak{}med.\allowbreak{}373294004.\allowbreak{}proc\_\allowbreak{}med\_\allowbreak{}admin & Medication/Event Flag & 25 & 100.0\% & 
100.0\% & 100.0\% & 0.0\% \\
152 & proc\_\allowbreak{}med.\allowbreak{}613391.\allowbreak{}proc\_\allowbreak{}med\_\allowbreak{}admin & Medication/Event Flag & 25 & 100.0\% & 
100.0\% & 100.0\% & 0.0\% \\
153 & proc\_\allowbreak{}med.\allowbreak{}1546356.\allowbreak{}proc\_\allowbreak{}med\_\allowbreak{}admin & Medication/Event Flag & 25 & 100.0\% & 
100.0\% & 100.0\% & 0.0\% \\
154 & race\_\allowbreak{}am\_\allowbreak{}indian & Administrative & 25 & 100.0\% & 100.0\% & 100.0\% & 0.0\% \\
155 & pt\_\allowbreak{}restriction2 & Administrative & 25 & 100.0\% & 100.0\% & 100.0\% & 0.0\% \\
156 & sex & Administrative & 25 & 100.0\% & 100.0\% & 100.0\% & 0.0\% \\
157 & stress\_\allowbreak{}performed & Binary Clinical Presence & 25 & 100.0\% & 100.0\% & 100.0\% & 0.0\% \\
\end{longtable}}
\end{document}